\documentclass{article}
\usepackage[preprint,eandd]{neurips_2026}

\usepackage[utf8]{inputenc}
\usepackage[T1]{fontenc}
\usepackage{hyperref}
\usepackage{url}
\usepackage{booktabs}
\usepackage{multirow}
\usepackage{amsfonts}
\usepackage{amsmath}
\usepackage{amssymb}
\usepackage{nicefrac}
\usepackage{microtype}
\usepackage[table]{xcolor}
\usepackage{graphicx}
\usepackage{subfig}

\definecolor{directbg}{RGB}{248,244,255}
\definecolor{implicitbg}{RGB}{242,248,242}
\definecolor{meanbg}{RGB}{245,245,245}
\newcommand{\Var}{\mathrm{Var}}

\title{A Tale of Two Variances: When Single-Seed Benchmarks Fail in Bayesian Deep Learning}

\author{
Qishi Zhan\thanks{Corresponding author.} \\
Department of Mathematical and Statistical Sciences \\
Marquette University, USA \\
\texttt{qishi.zhan@marquette.edu}
\And
Minxuan Hu \\
Cornell Ann S. Bowers College of Computing and Information Science \\
Cornell University, USA
\And
Liang He \\
Physics Science and Engineering \\
Tongji University, China
\And
Guansu Wang \\
School of Computing and Information Systems \\
The University of Melbourne, Australia
\And
Jiaxin Liu \\
Siebel School of Computing and Data Science \\
University of Illinois Urbana-Champaign, USA
}

\begin{document}

\maketitle

\begin{abstract}
In limited-data settings, a single endpoint mean of an evaluation metric such as the Continuous Ranked Probability Score (CRPS) is itself a random variable, yet it is routinely reported as if it were a stable property of the method. We study when this practice fails. Using 50 independent repetitions across six regression datasets, we show that CRPS variance trajectories differ substantially across methods and are not always well described by a smooth power-law decay. Methods with a learned heteroscedastic variance head, namely MAP and Deep Ensembles, can develop pronounced, reproducible variance peaks at intermediate training sizes on real datasets, whereas MC Dropout and Bayes by Backprop typically show smooth variance contraction. These peaks have direct practical consequences: at the variance peak on Seoul Bike, the relative RMSE of a single-seed MAP estimate reaches 93.6\%, and the probability of falling within \(\pm 10\%\) of the repeated-run mean drops to 5.9\%. We show that local CRPS variance provides a direct signal of single-seed estimation error, with Spearman correlations above 0.96 on every real dataset. Power-law fit quality and monotonicity together provide compact method-level 
summaries of trajectory regularity. Finally, replacing the standard heteroscedastic objective with \(\beta\)-NLL substantially reduces the irregular behavior, consistent with the view that the heteroscedastic training objective contributes to the instability. Practitioners should report trajectory summaries alongside endpoint means and concentrate repeated evaluation in high-variance regions.
\end{abstract}

\section{Introduction}

Bayesian deep learning benchmarks are usually reported through endpoint summaries of predictive performance under a fixed evaluation protocol. In probabilistic regression, a common choice is the Continuous Ranked Probability Score \citep{gneiting2007strictly}, or CRPS, reported as a single mean value on a held-out test set. This practice produces clean benchmark tables, but it can also encourage a stronger interpretation than the data support. A reported endpoint mean CRPS is often treated as a stable property of a method, even though it is the outcome of a stochastic training and evaluation pipeline \citep{bouthillier2021accounting,madaan2024variance,dehghani2021benchmark,reuel2024betterbench}.

This issue matters in limited-data settings, where repeated runs can lead to visibly different predictive distributions under the same architecture and test protocol. Prior work in Bayesian deep learning has focused mainly on endpoint quality, such as calibration, sharpness, and robustness \citep{lakshminarayanan2017simple,ovadia2019can,gustafsson2020evaluating,seligmann2023beyond,kirchhof2023url,mucsanyi2024benchmarking,gawlikowski2023survey,wilson2020bayesian}. More broadly, recent work has argued that benchmark results should be viewed as random quantities rather than fixed numbers \citep{bouthillier2021accounting,madaan2024variance,reuel2024betterbench,longjohn2025statistical}. We follow this view and ask a practical question. When does a single reported endpoint CRPS provide a reliable estimate of a method's expected performance, and when does it fail?

To answer this question, we study how the variance of CRPS changes with training set size \(n\). We find that this variance can become highly irregular across training sizes, with pronounced local peaks on several real datasets. These peaks matter because they mark regions where a single-seed endpoint report can become a poor estimate of the repeated-run mean. On Kin8nm, the relative RMSE of a single-seed CRPS estimate for MAP reaches 40.4\% at the variance peak, while the probability that a single-seed result falls within \(\pm 10\%\) of the repeated-run mean drops to 14.0\%. On Seoul Bike, the corresponding values are 93.6\% and 5.9\%. This suggests that the central issue is not only that some methods exhibit larger evaluation variance. In certain sample-size regimes, a standard one-run benchmark can substantially mismeasure the quantity it is intended to report.

This motivates a two-level view of benchmark reliability. At the training-size level, local CRPS variance captures the immediate risk of one-run mismeasurement. At the method level, power-law fit quality and monotonicity summarize whether evaluation behavior is globally regular or irregular across training sizes. In practice, these two levels serve different purposes: trajectory summaries show which method--dataset combinations deserve closer inspection, while local variance shows where one-run endpoint reports are most likely to fail.

Finally, we study one plausible source of the instability in direct variance learning \citep{nix1994estimating,kendall2017uncertainties,seitzer2022pitfalls}. Replacing the standard heteroscedastic negative log-likelihood with \(\beta\)-NLL substantially reduces the irregular behavior in the direct methods. This does not establish a complete mechanism, but it is consistent with the view that the training objective contributes to the dangerous sample-size regions seen in evaluation \citep{seitzer2022pitfalls}.

Our contributions are threefold. First, we show that Bayesian deep learning benchmarks can contain method-dependent training-size regions in which a single endpoint CRPS report becomes unreliable. Second, we show that benchmark unreliability is structured across training sizes, and that this structure is diagnostically useful in two different ways: power-law fit quality together with monotonicity provides helpful method-level summaries of trajectory regularity, while local variance serves as the direct risk measure at the training-size level. Third, we provide intervention-based evidence that the instability observed in direct methods is linked to the standard heteroscedastic training objective. Taken together, these results suggest that Bayesian deep learning benchmarks should report stability information alongside endpoint means, and should use more repetitions in regions where single-seed estimates are unreliable.


\section{Related Work}

Recent work has emphasized that benchmark results should be treated as random quantities rather than fixed numbers. \citet{bouthillier2021accounting} analyze several sources of variation in machine learning experiments, including data sampling, random initialization, and hyperparameter choices. \citet{madaan2024variance} study seed variability and monotonicity in evaluation pipelines for large models. These works show that single reported results can hide substantial uncertainty \citep{dehghani2021benchmark,raji2021everything,reuel2024betterbench,longjohn2024repositories,longjohn2025statistical}. Our work is aligned with this view, but focuses on a different question. We study when a single endpoint result is a poor estimate of expected performance, and how this failure depends on training set size and method.

Empirical studies of Bayesian deep learning have mainly focused on endpoint evaluation. Deep Ensembles \citep{lakshminarayanan2017simple} provide a strong baseline for predictive uncertainty, and later work studies calibration, robustness, and behavior under distribution shift \citep{ovadia2019can,gustafsson2020evaluating,seligmann2023beyond,ashukha2020pitfalls,gawlikowski2023survey,wilson2020bayesian}. These studies compare methods through final predictive performance under a fixed protocol. Our work instead studies the reliability of that reported endpoint itself. We ask how the variance of an evaluation metric changes with training set size, and whether this can identify regions where one-run evaluation is unreliable.

Our work is also related to scaling-law studies in deep learning. Early work showed that predictive loss and generalization often follow simple scaling patterns as data, model size, and compute increase \citep{kaplan2020scaling,hoffmann2022training}. Later work also examined deviations from ideal scaling behavior \citep{caballero2023broken}. In these studies, the main object is predictive accuracy or loss. We instead study the scaling behavior of evaluation uncertainty. This shift in focus matters because a method can show regular scaling in predictive performance while still showing irregular scaling in the variance of the metric used to evaluate it.

A related line of work studies scaling in uncertainty estimation itself. For example, \citet{rosso2026scaling} study how predictive uncertainty changes with data and model size. Their focus is on the uncertainty produced by the model. Our focus is different. We study the variability of the evaluation metric across repeated runs. This distinction matters because benchmark instability can arise even when the predictive uncertainty output appears well behaved.

Finally, our analysis connects to work on optimization pathologies in heteroscedastic regression. \citet{seitzer2022pitfalls} show that standard heteroscedastic negative log-likelihood training can lead to unstable optimization and propose a modified objective. We build on this result in a different way. Rather than studying prediction quality directly, we examine whether this training instability appears at the level of benchmark reliability. Our results suggest that the same pathology can contribute to dangerous sample-size regions in endpoint evaluation \citep{nix1994estimating,kendall2017uncertainties}.

\section{Evaluation Framework}
\label{sec:method}

We consider a regression task with input space $\mathcal{X}$ and output space $\mathcal{Y} \subseteq \mathbb{R}$. A Bayesian deep learning method $\mathcal{M}$ trained on a dataset of size $n$ produces a predictive distribution $p(y \mid x, \mathcal{D}_n)$ for each test input $x$. Given a held-out test set $\mathcal{D}_{\mathrm{test}}$ of size $N_{\mathrm{test}}$, we evaluate the method using the mean CRPS \citep{gneiting2007strictly,matheson1976scoring}
\begin{equation}
\widehat{\mathrm{CRPS}}_n
=
\frac{1}{N_{\mathrm{test}}}
\sum_{i=1}^{N_{\mathrm{test}}}
\mathrm{CRPS}\!\left(p(y \mid x_i, \mathcal{D}_n), y_i\right).
\label{eq:crps}
\end{equation}

Under a fixed benchmark protocol, $\widehat{\mathrm{CRPS}}_n$ is often reported as a single endpoint number. In our setting, however, it is a random variable. Its value depends on the sampled training set and on the stochastic training procedure. Our goal is to study the reliability of this reported endpoint. In particular, we ask when a single realization of $\widehat{\mathrm{CRPS}}_n$ is a reasonable estimate of the repeated-run mean, and when it is not.

\subsection{Variance and Single-Seed Reliability}
\label{sec:hypothesis}

Our analysis starts from the variance of the endpoint metric across repeated runs,
\begin{equation}
\mathrm{Var}\!\left[\widehat{\mathrm{CRPS}}_n\right].
\end{equation}
We study how this quantity changes with training set size $n$, and how its trajectory differs across methods. This variance is important for two reasons. First, it summarizes the stability of the evaluation pipeline at a given training size. Second, it determines how far a single-seed report can deviate from the repeated-run mean.

To describe the trajectory, we consider the approximate power-law form \citep{kaplan2020scaling,hoffmann2022training,caballero2023broken}
\begin{equation}
\mathrm{Var}\!\left[\widehat{\mathrm{CRPS}}_n\right]
\approx C n^{-\alpha},
\label{eq:powerlaw}
\end{equation}
where $C > 0$ and $\alpha > 0$ depend on the method and dataset. We do not treat this as a strict generative law. Instead, we use it as a descriptive model for whether the variance trajectory is regular or irregular across training sizes.

To clarify the source of this variability, we apply the law of total variance. Let $\theta$ denote the trained model parameters, which vary across runs because of stochastic training set sampling and random initialization. Then
\begin{equation}
\mathrm{Var}\!\left[\widehat{\mathrm{CRPS}}_n\right]
=
\underbrace{
\mathbb{E}\!\left[
\mathrm{Var}\!\left[\widehat{\mathrm{CRPS}}_n \mid \theta\right]
\right]
}_{\text{evaluation noise}}
+
\underbrace{
\mathrm{Var}\!\left[
\mathbb{E}\!\left[\widehat{\mathrm{CRPS}}_n \mid \theta\right]
\right]
}_{B_n}.
\label{eq:decomp}
\end{equation}
The first term reflects residual variation under a fixed trained model. The second term, $B_n$, captures variation across training runs. In our setting, the main differences in trajectory shape are driven by this second term. When $B_n$ is small, different runs lead to similar predictive distributions and similar endpoint CRPS values. When $B_n$ is large, single-seed reports become less reliable because different runs can end in substantially different solutions.

This decomposition suggests a simple interpretation. A smooth decrease in variance indicates that endpoint reliability improves steadily with training set size. A non-monotone spike, however, indicates a dangerous region in $n$, where a one-run benchmark report can become less reliable even as more training data are added. Our analysis therefore focuses on the shape of the variance trajectory, because reliability depends not only on how large the variance is overall, but also on whether it develops localized reversals across training sizes.

\subsection{Method-level trajectory summaries}
\label{sec:estimation}

We summarize each variance trajectory using two method-level quantities. The first is power-law fit quality. Given empirical variances $\hat{v}_n$ computed from $R=50$ independent realizations at each training size, we fit
\begin{equation}
\log \hat{v}_n = \log C - \alpha \log n + \varepsilon,
\label{eq:loglog}
\end{equation}
by ordinary least squares in log-log space. We report the coefficient of determination $R^2$ as a descriptive measure of how well the trajectory is captured by a single scaling pattern.

The second quantity is monotonicity. We classify a trajectory as monotone when the empirical variance is non-increasing with $n$, except for the boundary case where the maximum occurs at the smallest training size. This exception reflects the expected small-sample regime rather than the intermediate-size irregularity that is central to our study.

Together, \(R^2\) and monotonicity provide compact summaries of trajectory regularity for each method on each dataset. We do not use either quantity as a strict decision rule. Instead, they tell us whether a method--dataset combination looks globally regular or irregular across training sizes. They do not determine the exact training sizes where one-run evaluation becomes unreliable. For that, we examine the local variance values themselves.



\subsection{Datasets}

We evaluate on six regression datasets spanning synthetic and real-world settings, summarized in Table~\ref{tab:datasets}. The synthetic dataset provides a controlled setting with a known data-generating process. The real-world datasets cover distinct domains including robot kinematics, biological structure prediction, and urban energy use.

\subsection{Methods}

We group methods according to whether predictive uncertainty is produced through a learned heteroscedastic output head, since this distinction is central to the instability studied in this paper.

Direct parameterization methods learn predictive variance through a heteroscedastic Gaussian likelihood objective \citep{nix1994estimating,kendall2017uncertainties,seitzer2022pitfalls}. This group includes MAP and Deep Ensembles \citep{lakshminarayanan2017simple}. MAP trains a single network by maximizing the likelihood. Deep Ensembles trains \(M=5\) independent networks and combines their predictions, with each member sharing the same heteroscedastic output structure.

Implicit uncertainty methods obtain predictive uncertainty through stochastic averaging or posterior sampling, without a dedicated heteroscedastic variance head. MC Dropout (MCD) \citep{gal2016dropout} approximates posterior inference by performing \(T=50\) stochastic forward passes with dropout at test time. Bayes by Backprop (BBB) \citep{blundell2015weight} maintains a mean-field variational distribution \citep{blei2017variational} over network weights and draws \(T=50\) weight samples at test time. SWAG \citep{maddox2019simple} approximates the posterior in weight space using trajectories collected during training, but is excluded from the main trajectory analysis because it often fails to collect sufficient weight snapshots for reliable posterior approximation at small \(n\). Full implementation details are given in Appendix~\ref{app:implement_details}.


\section{Results}

\subsection{Evaluation Protocol}

For each dataset, we fix a held-out test set comprising $30\%$ of the total samples. For each training size $n \in \{10,20,30,50,100,200,500,1000,2000\}$ and each method, we draw $R=50$ independent training sets of size $n$ from the remaining pool and record $\widehat{\mathrm{CRPS}}_n$ for each realization. We compute the empirical variance $\hat{v}_n$ from these repeated runs and use it to study both variance trajectories and single-seed reliability.

To summarize trajectory shape, we fit the log-log model in Equation~\ref{eq:loglog} by ordinary least squares and report $\alpha$, $C$, and $R^2$. We also record whether the empirical variance trajectory is monotone. Together, these quantities provide compact summaries of trajectory regularity used in our analysis.

Whenever model selection is required, it is performed using a validation split. The held-out test set is used only for final evaluation. SWAG is excluded from the main trajectory analysis for the reasons described in Section~3.4.

\subsection{Variance trajectories reveal stable intermediate-\(n\) spikes}

\begin{figure}[t]
\centering
\includegraphics[width=0.8\linewidth]{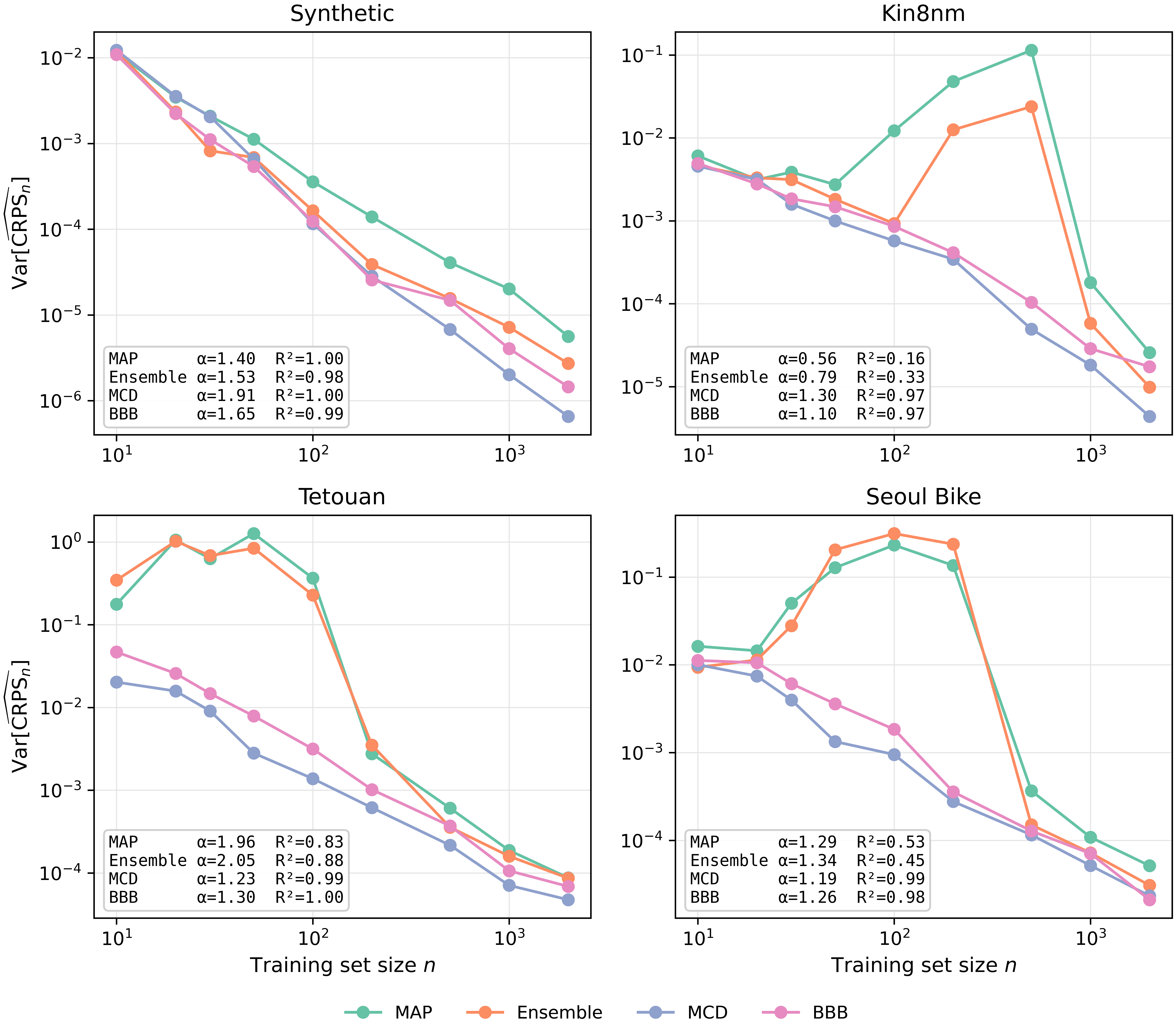}
\caption{CRPS variance trajectories across four representative datasets, with fitted power-law curves $Cn^{-\alpha}$. MCD and BBB show smooth variance contraction. MAP and Deep Ensembles exhibit non-monotone spikes at intermediate training sizes.}
\label{fig:traj}
\label{fig:scaling}
\end{figure}

With this protocol in place, we now examine how variance trajectories differ across methods.
Figure~\ref{fig:scaling} shows that some method--dataset combinations exhibit clear local variance peaks at intermediate training sizes, rather than a smooth decrease in variance as \(n\) increases. This pattern is most visible for MAP and Deep Ensembles on Kin8nm, Tetouan, and Seoul Bike. By contrast, MCD and BBB usually show smoother variance contraction across the same range of training sizes.

These spikes are not visible on every dataset. On the real datasets, Make Regression is an exception: the family-level split between direct and implicit methods is less clear there. This is consistent with a near-linear signal structure that may make optimization less sensitive to initialization; full details are in Appendix~\ref{app:make}. The main exception among the implicit methods is BBB on Protein, where the trajectory is non-monotone because of two extreme spikes at \(n=20\) and \(n=200\). These are better understood as episodic divergence events than as a broad failure of regular scaling. Outside these two points, the trajectory is much more regular.

\begin{figure}[t]
\centering
\includegraphics[width=\linewidth]{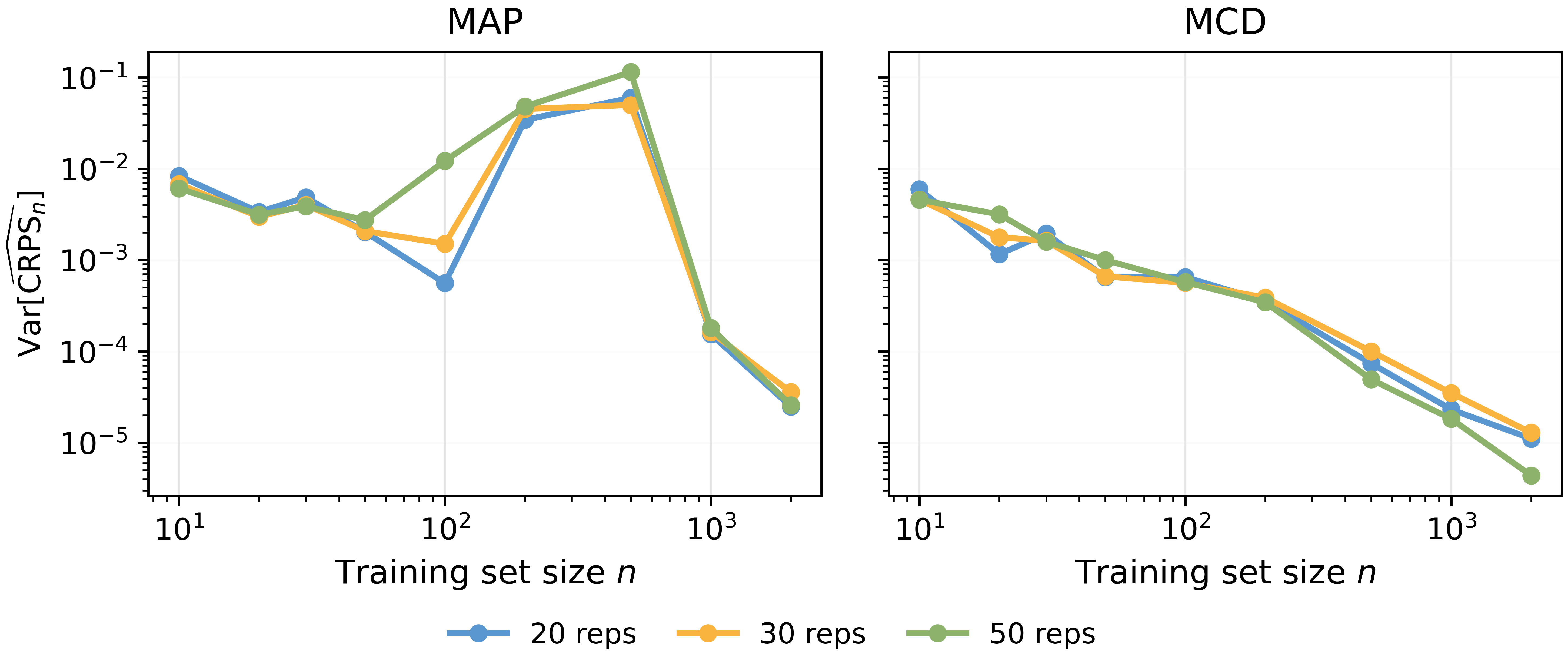}
\caption{Variance trajectories on Kin8nm under $R = 20, 30, 50$ for all four methods. 
The MAP and Ensemble spikes remain centered at $n = 500$ across all three repetition 
counts, while MCD and BBB remain smooth throughout, confirming that the spikes reflect 
stable features of method behavior rather than finite-sample fluctuations.}
\label{fig:r_stability}
\end{figure}

A natural question is whether the intermediate-\(n\) spikes could be artifacts of using \(R=50\) repetitions. To examine this, Figure~\ref{fig:r_stability} compares variance trajectories on Kin8nm under \(R \in \{20,30,50\}\). If the spike were only a finite-sample fluctuation, its location or relative prominence would be expected to vary substantially across choices of \(R\). Instead, the MAP and Ensemble spikes remain centered at $n = 500$ under all three 
repetition counts, while MCD and BBB remain smooth throughout. This supports the view 
that the spike reflects a stable feature of direct method behavior at that training 
size rather than an accidental fluctuation caused by one particular choice of $R$.

Table~\ref{tab:powerlaw} summarizes these trajectories through power-law fit quality and monotonicity. We use these quantities only as descriptive method-level summaries. They do not establish the existence of a spike on their own, but they provide a compact way to compare whether a trajectory is globally regular or irregular across method--dataset combinations.

A clear difference appears between the two method groups. The implicit methods usually show smooth variance contraction as \(n\) increases and achieve high \(R^2\) on most datasets. MCD achieves high \(R^2\) on five of six datasets, with values ranging from 0.894 
to 0.996, and BBB shows similarly regular trajectories on five of six datasets. 
These methods are also usually monotone. In contrast, the direct methods show much weaker fit quality on the real datasets. MAP and Deep Ensembles both exhibit low \(R^2\) on Kin8nm, Tetouan, and Seoul Bike, and both are non-monotone on these datasets.

On the synthetic dataset, all four methods exhibit monotone trajectories and high \(R^2\) values, with no evidence of intermediate-\(n\) variance spikes. This is consistent with the controlled heteroscedastic structure of the synthetic data-generating process, which does not produce the kind of irregular learning dynamics seen on real datasets. This contrast suggests that the instability observed on real datasets is associated with the structure of real-world data rather than being an inherent property of the methods themselves.

Taken together, Figure~\ref{fig:scaling} and Figure~\ref{fig:r_stability} show that the intermediate-\(n\) spikes are stable features of evaluation behavior in several direct methods on real datasets rather than artifacts of one particular choice of repetition count. The next subsection quantifies how much practical error these spikes create in one-run benchmark reporting.

\begin{table}[t]
\centering
\small
\setlength{\tabcolsep}{5pt}
\renewcommand{\arraystretch}{1.15}
\caption{Power-law fit quality and monotonicity for each method and dataset. Entries report \(R^2\) and monotonicity status (M: monotone, NM: non-monotone). These quantities are used as descriptive summaries of trajectory regularity rather than as strict decision rules.}
\label{tab:powerlaw}
\begin{tabular}{llcccccc}
\toprule
Type & Method & Synthetic & Kin8nm & Protein & Make & Tetouan & Seoul Bike \\
\midrule
\rowcolor{directbg}
Direct
& MAP
& 0.998 / M
& 0.158 / NM
& 0.882 / NM
& 0.838 / NM
& 0.825 / NM
& 0.530 / NM \\
\rowcolor{directbg}
& Ensemble
& 0.984 / M
& 0.335 / NM
& 0.867 / NM
& 0.878 / NM
& 0.879 / NM
& 0.453 / NM \\
\midrule
\rowcolor{implicitbg}
Implicit
& MCD
& 0.996 / M
& 0.969 / M
& 0.894 / M
& 0.976 / M
& 0.989 / M
& 0.991 / M \\
\rowcolor{implicitbg}
& BBB
& 0.988 / M
& 0.975 / M
& 0.338 / NM
& 0.957 / M
& 0.995 / M
& 0.979 / M \\
\midrule
\rowcolor{meanbg}
\multicolumn{2}{l}{Implicit mean \(R^2\)} & 0.992 & 0.972 & 0.616 & 0.967 & 0.992 & 0.985 \\
\rowcolor{meanbg}
\multicolumn{2}{l}{Direct mean \(R^2\)} & 0.991 & 0.247 & 0.875 & 0.858 & 0.852 & 0.492 \\
\bottomrule
\end{tabular}
\end{table}


\subsection{Single-seed endpoint reports fail in high-variance regions}

The variance patterns identified above are not only descriptive features of the trajectory. They have a direct consequence for benchmark reliability. When the local variance of CRPS is large at a given training size, a single-seed endpoint report can be a poor estimate of the repeated-run mean.

To quantify this effect, we study the self-estimation error of one-run evaluation. For each method and training size, we repeatedly sample one run from the 50 repeated trials and compare its CRPS to the repeated-run mean. We summarize the discrepancy by the relative RMSE and by the probability that a single-seed result falls within $\pm 10\%$ of the repeated-run mean.

The practical consequences are clearest for MAP on datasets where local variance becomes large at intermediate training sizes. On Kin8nm, the relative RMSE of a single-seed CRPS estimate reaches $40.4\%$ at $n=500$, which is also the variance peak. At this point, the probability that a single-seed result falls within $\pm10\%$ of the repeated-run mean is only $14.0\%$. On Seoul Bike, the pattern is more pronounced: relative RMSE reaches $93.6\%$ at $n=100$, and the probability of falling within $\pm10\%$ drops to $5.9\%$. On Tetouan, MAP variance peaks at $n=20$, where the relative RMSE reaches $77.3\%$ and the probability of falling within $\pm10\%$ is effectively $0\%$. In all three cases, a single-seed endpoint report at the variance peak is a poor estimate of the repeated-run mean for the large majority of runs.

By contrast, MCD remains much more stable across training sizes. Its relative RMSE stays below $10\%$ on Kin8nm and below $20\%$ on both Seoul Bike and Tetouan, with the worst values occurring at the smallest training sizes rather than at intermediate-$n$ variance peaks. This contrast shows that the practical issue is not only that some methods have larger evaluation variance on average. The issue is that large local variance can induce substantial measurement error in one-run endpoint evaluation, and this risk is concentrated at specific training sizes rather than spread uniformly across the data-scale axis.

This pattern is also visible in the full peak-error summary across methods and datasets. Table~\ref{tab:single_seed} shows that the largest single-seed errors for direct methods are typically much larger than those for implicit methods on the main real datasets, with Make Regression and BBB on Protein providing the main exceptions. For BBB on Protein, the extreme $560.5\%$ value reflects episodic variational-inference divergence at $n=200$ rather than systematic instability; see Appendix~\ref{app:bbb_protein}. These peak-error comparisons are practically important because they show how badly a benchmark report can be distorted at the worst training size for each method.

Beyond quantifying the magnitude of these failures, local variance also provides a consistent signal of where they occur. Across method-$n$ pairs, the Spearman correlation between $\Var[\widehat{\mathrm{CRPS}}_n]$ and relative RMSE ranges from $0.966$ to $0.998$ across all five real datasets, with $p$-values below $10^{-20}$ in every case. When all datasets are pooled, the correlation remains $0.904$ ($p = 1.8 \times 10^{-67}$). A fixed-effects robustness analysis controlling for average differences across datasets and methods gives the same qualitative conclusion. Dataset-wise correlations are reported in Appendix~\ref{app:corr}, and the robustness analysis is reported in Appendix~\ref{app:fe_robustness}. Together, these results indicate that local variance is a reliable signal of single-seed estimation error across methods and datasets.

The same conclusion appears when method-$n$ pairs are grouped by local variance rather than by method label. On every real dataset, the top quartile of local variance has substantially larger single-seed error than the remaining three quartiles, and relative RMSE increases monotonically across quartiles without exception. On Kin8nm, the mean relative RMSE in the top-quartile region is $20.2\%$, compared to $5.5\%$ in the remaining points (ratio $3.7\times$, Mann--Whitney $p = 4.9 \times 10^{-6}$). On Seoul Bike, the corresponding values are $65.1\%$ versus $12.9\%$ ($5.0\times$, $p = 5.8 \times 10^{-6}$). On Tetouan, they are $62.7\%$ versus $9.0\%$ ($7.0\times$, $p = 4.9 \times 10^{-6}$). Full results for all five datasets, including quartile-level breakdowns and Mann–Whitney test statistics, are reported in Appendix~\ref{app:highvar_full}.

Taken together, these results show that local variance has immediate practical meaning. In high-variance regions of $n$, a standard single-seed endpoint report can be a poor estimate of the quantity that the benchmark is intended to summarize. This is therefore not only a comparison problem between methods, but a measurement problem in the benchmark itself. Full results for all methods and datasets are reported in Appendix~\ref{app:single_seed}. We next ask what drives these spike regions.

\begin{figure}[t]
\centering
\includegraphics[width=\linewidth]{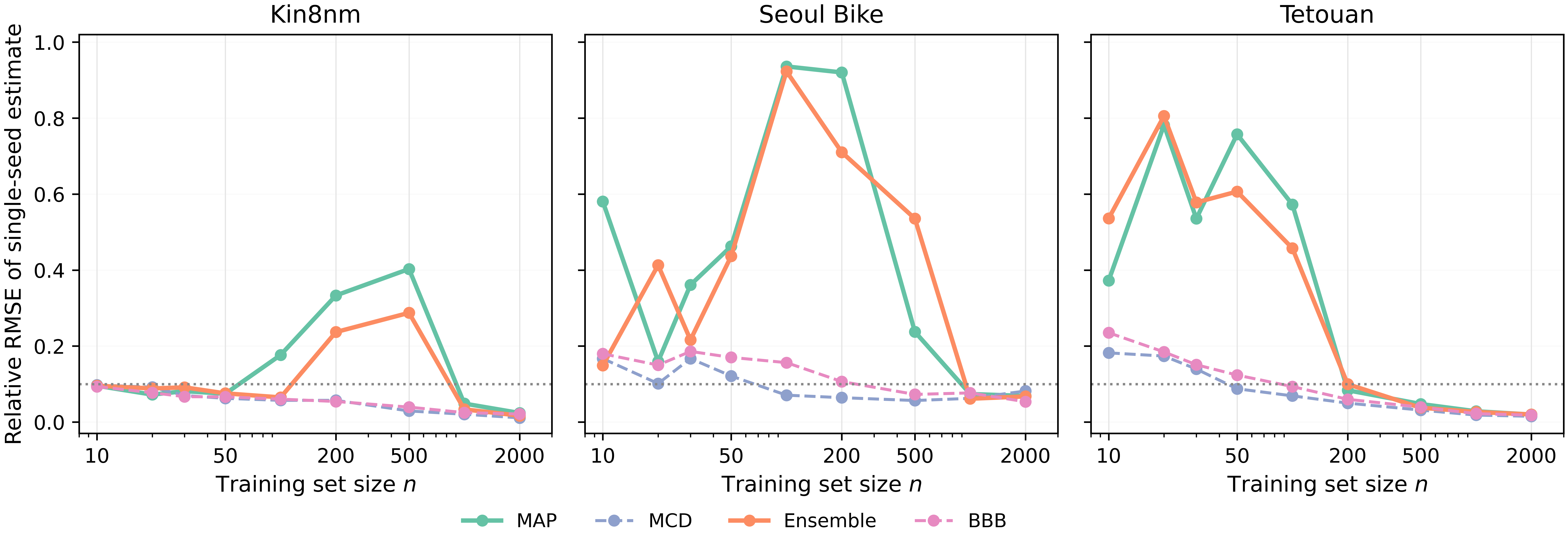}
\caption{Single-seed estimation error across training sizes on representative real datasets. Direct methods develop pronounced error peaks where local CRPS variance is largest; MCD and BBB remain stable. Dotted line marks the $10\%$ reference level.}
\label{fig:seed}
\end{figure}

\subsection{\(\beta\)-NLL reduces instability in direct variance learning}


The concentration of instability at intermediate training sizes, rather than at the smallest 
or largest $n$, suggests that this regime has a structurally distinct character. Prior work 
on interpolation thresholds and double descent has shown that model behavior becomes 
especially sensitive to initialization when effective model complexity and data size are 
poorly matched \citep{belkin2019reconciling, nakkiran2020deep, hastie2022surprises}: in 
this critical regime, different random initializations can follow substantially different 
optimization trajectories, producing larger run-to-run variability than in either the 
clearly underfit or clearly data-rich settings. One plausible explanation is that 
intermediate training sizes place direct parameterization methods in precisely this 
sensitive regime, where the additional gradient-attenuation effect of the heteroscedastic 
objective \citep{seitzer2022pitfalls} can then amplify the resulting instability into the 
pronounced variance spikes observed in our experiments. A fuller discussion is given in 
Appendix~\ref{app:mechanism}.

We next examine whether the standard heteroscedastic negative log-likelihood contributes to the irregular variance trajectories observed above.

\begin{figure}[htbp]
\centering
\begin{minipage}[c]{0.45\linewidth}
    \centering
    \includegraphics[width=\linewidth]{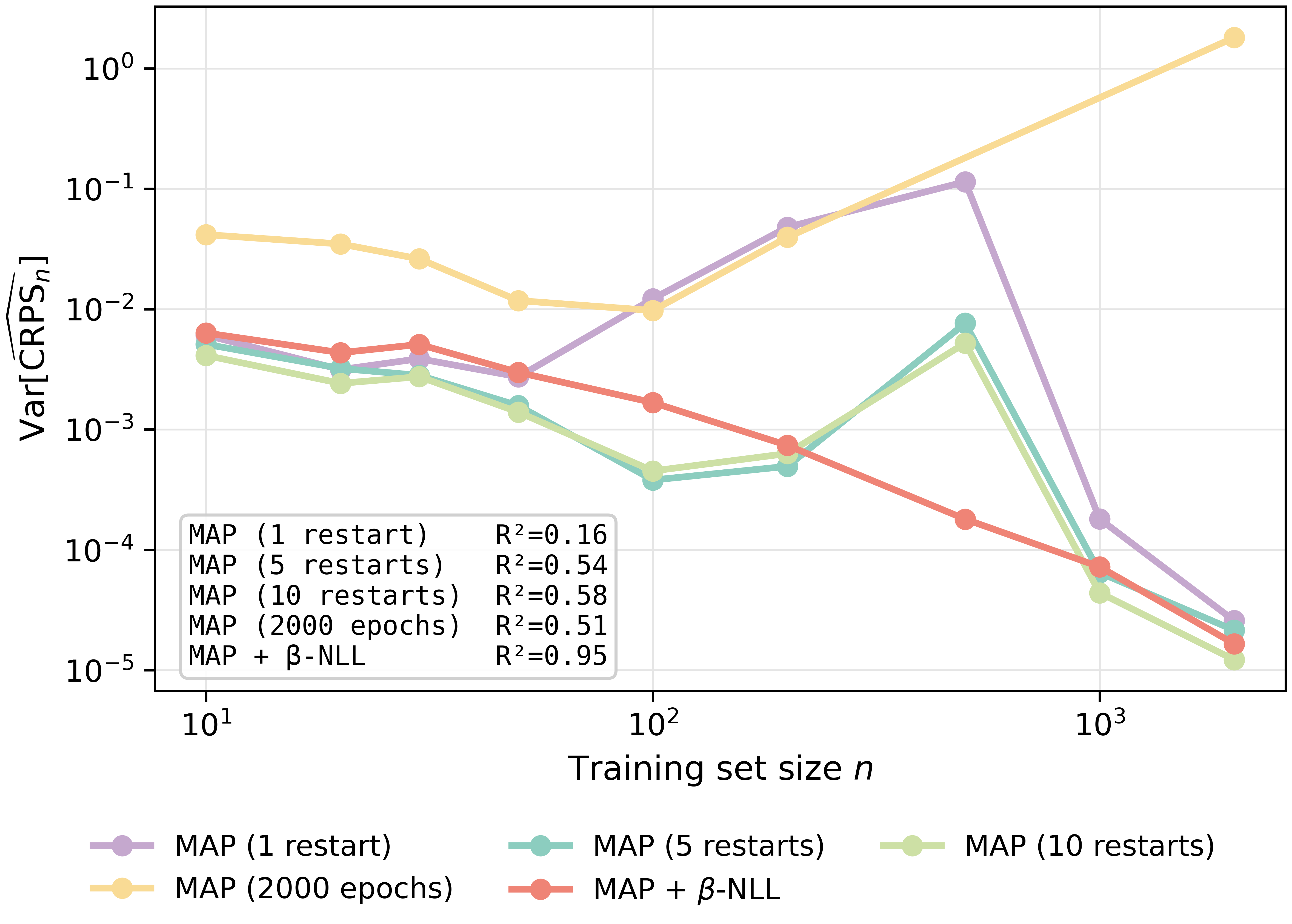}
    \caption{CRPS variance trajectories on Kin8nm for several MAP-based interventions. Additional restarts and longer training improve the trajectory only partially. MAP trained with $\beta$-NLL produces a much more regular contraction pattern.}
    \label{fig:causal}
\end{minipage}
\hfill
\begin{minipage}[c]{0.50\linewidth}
    \centering
    \captionof{table}{Stability signatures for MAP-based intervention variants.}
    \label{tab:intervention}
    \small
    \begin{tabular}{llcc}
    \toprule
    Dataset & Method & $R^2$ & Monotone \\
    \midrule
    \multirow{5}{*}{Kin8nm}
    & MAP (baseline) & 0.158 & No \\
    & MAP5           & 0.543 & No \\
    & MAP10          & 0.583 & No \\
    & MAP (2000 ep.) & 0.511 & No \\
    & MAP+$\beta$-NLL & 0.950 & Yes \\
    \midrule
    \multirow{2}{*}{Tetouan}
    & MAP (baseline)  & 0.825 & No \\
    & MAP+$\beta$-NLL & 0.989 & Yes \\
    \midrule
    \multirow{2}{*}{Seoul Bike}
    & MAP (baseline)  & 0.530 & No \\
    & MAP+$\beta$-NLL & 0.976 & Yes \\
    \bottomrule
    \end{tabular}
\end{minipage}
\end{figure}

We focus on Kin8nm, where the instability is most pronounced, and compare several 
MAP-based interventions (Table~\ref{tab:intervention}). Additional random restarts 
improve the trajectory only partially. Additional random restarts improve the trajectory only partially. MAP5 increases $R^2$ from $0.158$ to $0.543$, and MAP10 reaches $0.583$, but both remain non-monotone. Training for 2000 epochs also leaves the instability largely unchanged, with $R^2=0.511$.

In contrast, replacing the standard objective with \(\beta\)-NLL produces a much more 
regular variance trajectory, with \(R^2=0.950\) on Kin8nm (Figure~\ref{fig:causal}). 
The same pattern appears on Tetouan and Seoul Bike, where MAP+\(\beta\)-NLL reaches 
\(R^2=0.989\) and \(0.976\). These results do not prove a complete mechanism, but they 
are consistent with the view that the training objective contributes to the dangerous 
sample-size regions observed in endpoint evaluation. This interpretation is further supported by the classification experiments in Appendix~\ref{app:classification}, where intermediate-\(n\) spikes are absent across all four datasets under Brier score, and no family-level split appears between direct and implicit methods.

\section{Conclusion}

We have studied the reliability of single-seed endpoint evaluation in Bayesian deep 
learning benchmarks. Repeated-run experiments show that CRPS variance trajectories 
differ substantially across methods, and that direct variance learning methods on 
real-world data can develop stable localized spikes at intermediate training sizes. 
These spikes mark the regions where a one-run endpoint report is most likely to be 
a poor estimate of expected performance. Local CRPS variance is the most direct 
measure of risk at a specific training size, while trajectory summaries such as 
\(R^2\) and monotonicity describe whether a method exhibits globally regular or 
irregular evaluation behavior. A \(\beta\)-NLL intervention substantially reduces 
the instability in direct methods, and the absence of analogous spikes in 
classification experiments under Brier score provides additional support for the 
view that the heteroscedastic training objective is the primary contributing factor. 
These conclusions rest on a shared two-hidden-layer MLP setup across six regression 
datasets; whether analogous spike structures emerge under deeper architectures, 
alternative optimizers, or larger data regimes remains an open question. Among the 
regression metrics examined, CRPS and interval score provide the most consistent 
diagnostic signal, while negative log-likelihood and interval coverage probability 
are less reliable for this purpose.

Our results suggest three concrete actions for practitioners. First, benchmark reports 
should include trajectory summaries such as \(R^2\) and monotonicity alongside 
endpoint means, because they help flag method and dataset combinations that deserve 
closer inspection. Second, repeated evaluation should be used to identify the training 
sizes where local variance is largest, since that is where additional repetitions are 
most informative rather than being applied uniformly across all training sizes. Third, 
for direct variance learning methods, replacing the standard heteroscedastic objective 
with \(\beta\)-NLL is worth considering as a low-cost intervention that substantially 
regularizes evaluation behavior without requiring architectural changes.

More broadly, the central issue identified here is not specific to Bayesian deep 
learning. A reported endpoint mean is always the outcome of a stochastic pipeline, 
and in limited-data regimes it can be a poor summary of expected performance 
regardless of the method being evaluated. The diagnostic tools developed here, 
variance trajectories, power-law fit quality, and local variance as a risk signal, 
are applicable to any benchmark setting where repeated evaluation is feasible. 
Treating reported metrics as random quantities rather than fixed properties of a 
method would bring benchmark practice closer to the statistical realities of 
stochastic training pipelines.

\bibliographystyle{plainnat}
\bibliography{ref}

\appendix

\section{Experimental Setup}
\label{app:experiment}

\subsection{Datasets}
\label{app:datasets}

\begin{table}[h]
\centering
\caption{Datasets used in our empirical study.
Pool size is the number of samples available for training
after reserving 30\% for testing.}
\label{tab:datasets}
\begin{tabular}{lrrrp{5cm}}
\toprule
Dataset & Total & Pool & Features & Source \\
\midrule
Synthetic         & 5{,}000  & 3{,}500  & 8  & Custom heteroscedastic \\
Kin8nm            & 8{,}192  & 5{,}735  & 8  & UCI / OpenML \citep{ghahramani1996kin} \\
Protein Structure & 45{,}730 & 32{,}011 & 9  & UCI / OpenML \citep{rana2013protein} \\
Make Regression   & 20{,}000 & 14{,}000 & 10 & \texttt{sklearn} \citep{pedregosa2011scikit} \\
Tetouan     & 52{,}417 & 36{,}691 & 6  & UCI \citep{salam2018tetouan} \\
Seoul Bike        & 8{,}760  & 6{,}132  & 8  & UCI \citep{seoulbike2020dataset} \\
\bottomrule
\end{tabular}
\end{table}

\subsection{Implementation details}
\label{app:implement_details}

All methods share a common two-hidden-layer MLP architecture with 64 units
per layer and ReLU activations.
The output layer produces two scalars corresponding to the predicted mean
$\mu$ and log variance $\log\sigma^2$.
Predicted variance is clamped to $[10^{-3},10^3]$ during training for
numerical stability.

MAP, MCD, and Deep Ensembles are trained for 500 epochs using the Adam
optimizer \citep{kingma2015adam} with learning rate $10^{-3}$ and weight decay $10^{-5}$.
BBB uses 1000 epochs to allow variational convergence and omits weight
decay, since the KL divergence term already provides regularization.
For MCD, BBB, and SWAG we draw $T=50$ stochastic samples at test time.
All experiments were conducted on NVIDIA A100 GPUs.

\subsection{Single-seed self-estimation protocol}
\label{app:self_estimation}

For a fixed method and training size $n$, let
$\{\widehat{\mathrm{CRPS}}_n^{(r)}\}_{r=1}^{50}$
denote the 50 independent CRPS values.
We use their empirical mean $\bar{c}_n$ as the reference estimate of
expected performance.
We then sample one run with replacement 50{,}000 times and compute the
relative error of each draw,
\[
  \mathrm{RelErr}(c_n^\star)
  = \frac{c_n^\star - \bar{c}_n}{\bar{c}_n}.
\]
We summarize this distribution by the relative RMSE
\[
  \mathrm{rel\text{-}RMSE}_n
  = \sqrt{\mathbb{E}\!\left[\left(
      \frac{c_n^\star - \bar{c}_n}{\bar{c}_n}
    \right)^2\right]},
\]
and by the probability $P\!\left(|c_n^\star-\bar{c}_n|/\bar{c}_n \le 0.1\right)$.
Both quantities characterize practical benchmark reliability under the
observed repeated-run distribution and should not be interpreted as
exact population quantities.

\subsection{Theoretical motivation for intermediate-\(n\) instability}
\label{app:mechanism}

We outline one plausible explanation for why instability in direct parameterization methods can be concentrated at intermediate training sizes. This discussion is intended as theoretical motivation rather than a formal proof.

One possible factor is heightened sensitivity in the intermediate regime. Prior work on interpolation and double descent suggests that model behavior can become especially sensitive when effective model complexity and data constraints are poorly matched \citep{belkin2019reconciling,nakkiran2020deep,hastie2022surprises}. In such regimes, different random initializations may follow substantially different optimization paths, leading to larger run-to-run variability than in clearly underfit or clearly data-constrained settings.

A second factor comes from the heteroscedastic negative log-likelihood itself. The standard objective \citep{nix1994estimating,kendall2017uncertainties,seitzer2022pitfalls} is
\[
  \mathcal{L}(\mu,\sigma^2)
  = \tfrac{1}{2}\log\sigma^2
    + \frac{(y-\mu)^2}{2\sigma^2},
\]
with gradient
\[
  \frac{\partial\mathcal{L}}{\partial\mu}
  = -\frac{y-\mu}{\sigma^2}.
\]
As emphasized by \citet{seitzer2022pitfalls}, this scaling can attenuate the gradient available to correct the mean prediction when the model inflates \(\sigma^2\). In unfavorable runs, this may create a self-reinforcing pattern in which poor mean estimates are accompanied by large variance estimates, slowing further correction.

A interpretation is that intermediate training sizes are the regime where these two effects interact most strongly. The data are informative enough for different runs to separate, but not always strong enough to suppress unfavorable trajectories once gradient attenuation sets in. This can produce a small number of anomalous runs with unusually poor CRPS, thereby inflating empirical variance.

The \(\beta\)-NLL objective weakens the dependence of the mean gradient on the predicted variance \citep{seitzer2022pitfalls}. For \(\beta=0.5\), the attenuation scales as \(\sigma^{-1}\) rather than \(\sigma^{-2}\), which can make mean optimization less vulnerable to variance inflation. Our intervention results are consistent with this interpretation.

Methods such as MCD and BBB do not learn predictive uncertainty through the same direct heteroscedastic variance head used by MAP and Deep Ensembles. They are therefore less directly exposed to this particular mean-gradient attenuation mechanism, although they may still exhibit other forms of instability. The \(\beta\)-NLL intervention addresses the second factor more directly, but the relative contribution of the two mechanisms cannot be isolated from the current experiments.

\subsection{Valid-run patterns}
\label{app:valid_runs}

MAP, Deep Ensembles, MCD, and BBB produce valid outputs for all training
sizes on all datasets.
SWAG shows reduced valid-run rates in multiple small-$n$ regimes due to
insufficient weight snapshots for reliable posterior approximation;
it is therefore excluded from the main analysis.
Detailed valid-run counts are shown in Table~\ref{tab:valid_patterns}.

\begin{table}[h]
\centering
\small
\caption{Valid-run patterns across methods and datasets.}
\label{tab:valid_patterns}
\begin{tabular}{llp{7cm}}
\toprule
Method & Dataset & Valid-run pattern \\
\midrule
MAP, Ensemble, MCD, BBB & All datasets & Valid for all $n$ \\
\midrule
SWAG & Synthetic    & Reduced for $n\in\{10,20,30,50\}$; full from $n\ge100$ \\
SWAG & Kin8nm       & Reduced for $n\le100$; full from $n\ge200$ \\
SWAG & Protein      & Reduced for $n\in\{10,20,30,50\}$; full from $n\ge100$ \\
SWAG & Make         & Reduced for all $n$; no regime with full validity \\
SWAG & Tetouan      & Reduced for $n\in\{10,20,30\}$; full from $n\ge50$ \\
SWAG & Seoul Bike   & Reduced for $n\le200$; full from $n\ge500$ \\
\bottomrule
\end{tabular}
\end{table}

\section{Single-Seed Reliability: Full Results}
\label{app:single_seed}


Table~\ref{tab:single_seed} reports peak relative RMSE and
$P(\pm10\%)$ for all methods and datasets.
The three datasets used in the main text (Kin8nm, Seoul Bike, Tetouan)
show the clearest family-level split between direct and implicit methods.

\begin{table}[h]
\caption{Single-seed self-estimation error at the worst training size
for each method and dataset.
Direct methods (MAP, Ensemble) consistently show higher peak errors
than implicit methods (MCD, BBB), except on Make Regression
(see text).}
\label{tab:single_seed}
\centering
\small
\begin{tabular}{llrrr}
\toprule
Dataset & Method & Max rel-RMSE & Peak $n$ & $P(\pm10\%)$ \\
\midrule
\multirow{4}{*}{Kin8nm}
  & MAP      & $40.4\%$ & $500$ & $14.0\%$ \\
  & Ensemble & $28.7\%$ & $500$ & $25.9\%$ \\
  & MCD      & $9.7\%$  & $10$  & $72.1\%$ \\
  & BBB      & $9.3\%$  & $10$  & $68.2\%$ \\
\midrule
\multirow{4}{*}{Seoul Bike}
  & MAP      & $93.6\%$ & $100$ & $5.9\%$  \\
  & Ensemble & $91.4\%$ & $100$ & $16.4\%$ \\
  & MCD      & $16.7\%$ & $30$  & $65.8\%$ \\
  & BBB      & $18.6\%$ & $30$  & $55.7\%$ \\
\midrule
\multirow{4}{*}{Tetouan}
  & MAP      & $77.3\%$ & $20$  & $0.0\%$  \\
  & Ensemble & $80.3\%$ & $20$  & $12.0\%$ \\
  & MCD      & $18.1\%$ & $10$  & $36.2\%$ \\
  & BBB      & $23.6\%$ & $10$  & $27.9\%$ \\
\midrule
\multirow{4}{*}{Make Regression}
  & MAP      & $22.1\%$ & $10$  & $29.6\%$ \\
  & Ensemble & $25.1\%$ & $10$  & $28.1\%$ \\
  & MCD      & $26.4\%$ & $20$  & $25.9\%$ \\
  & BBB      & $25.1\%$ & $20$  & $29.9\%$ \\
\midrule
\multirow{4}{*}{Protein}
  & MAP      & $25.9\%$ & $10$  & $28.1\%$ \\
  & Ensemble & $39.4\%$ & $20$  & $26.2\%$ \\
  & MCD      & $17.6\%$ & $10$  & $38.0\%$ \\
  & BBB      & $560.5\%$& $200$ & $0.0\%$  \\
\bottomrule
\end{tabular}
\end{table}

\subsection{Make Regression Exception}
\label{app:make}

On Make Regression the family-level split between direct and
implicit methods does not appear clearly: MCD peaks at 26.4\%
relative RMSE, marginally exceeding MAP's 22.1\%.
All methods peak at small $n$ (10 or 20), consistent with
small-sample noise rather than the intermediate-$n$ spikes seen on
Kin8nm, Tetouan, and Seoul Bike.
The near-linear signal structure of this dataset (generated by
\texttt{sklearn.make\_regression} with low noise) likely makes the
heteroscedastic NLL objective less prone to the variance-inflation
pathology at intermediate $n$: when the signal is approximately
linear, the optimizer faces a more stable curvature landscape across
random initializations.
 
The local variance analysis in Appendix~\ref{app:corr} still shows a significant high-variance-region effect on Make Regression, even though the family-level split is absent.

\paragraph{BBB on Protein.}
The $560.5\%$ value reflects episodic variational inference divergence
at $n=200$ rather than systematic instability;
see Appendix~\ref{app:bbb_protein}.




\section{Local Variance as a Predictor of Single-Seed Error}
\label{app:local_var}


\subsection{Spearman Correlation: All Datasets}
\label{app:corr}

Table~\ref{tab:spearman_full} reports the Spearman correlation between
local $\mathrm{Var}[\widehat{\mathrm{CRPS}}_n]$ and relative RMSE
across all method-$n$ pairs on each dataset.
The correlation exceeds 0.96 on every dataset, including Make Regression where the family-level split is less clear \citep{demsar2006statistical}. This supports the use of local variance as a consistent risk measure at the training-size level even when the broader trajectory pattern is less informative.

\begin{table}[h]
\caption{Spearman correlation between local Var(CRPS) and
single-seed relative RMSE across all method-$n$ pairs.}
\label{tab:spearman_full}
\centering
\begin{tabular}{lrr}
\toprule
Dataset & Spearman $\rho$ & $p$-value \\
\midrule
Kin8nm          & $0.990$ & $1.8\times10^{-30}$ \\
Seoul Bike      & $0.979$ & $3.3\times10^{-25}$ \\
Make Regression & $0.966$ & $1.3\times10^{-21}$ \\
Protein         & $0.991$ & $1.7\times10^{-31}$ \\
Tetouan         & $0.998$ & $2.8\times10^{-44}$ \\
\midrule
All combined    & $0.904$ & $1.8\times10^{-67}$ \\
\bottomrule
\end{tabular}
\end{table}

\subsection{High-variance region vs remaining points}
\label{app:highvar_full}

We define the high-variance region on each dataset as the top quartile
of local $\mathrm{Var}[\widehat{\mathrm{CRPS}}_n]$ across all
method-$n$ pairs.
Table~\ref{tab:highvar} compares mean relative RMSE in the
high-variance region against the remaining three quartiles.
The ratio ranges from $2.6\times$ to $13.4\times$, and the difference
is significant under a one-sided Mann-Whitney test on every dataset \citep{demsar2006statistical}.
Relative RMSE increases monotonically across all four quartiles on
every real dataset, confirming the result is not driven by outliers.

\begin{table}[h]
\caption{Mean single-seed relative RMSE in the top-quartile
variance region versus the remaining points.}
\label{tab:highvar}
\centering
\begin{tabular}{lrrrr}
\toprule
Dataset & High-var & Rest & Ratio & MW $p$ \\
\midrule
Kin8nm          & $20.2\%$ & $5.5\%$  & $3.7\times$ & $4.9\times10^{-6}$ \\
Seoul Bike      & $65.1\%$ & $12.9\%$ & $5.0\times$ & $5.8\times10^{-6}$ \\
Make Regression & $23.2\%$ & $9.0\%$  & $2.6\times$ & $9.6\times10^{-6}$ \\
Protein$^\dagger$ & $105.4\%$ & $7.8\%$ & $13.4\times$ & $5.8\times10^{-6}$ \\
Tetouan         & $62.7\%$ & $9.0\%$  & $7.0\times$ & $4.9\times10^{-6}$ \\
\bottomrule
\end{tabular}
\smallskip

{\small $^\dagger$ Elevated by the BBB episodic divergence at $n=200$.
Excluding that point reduces the high-var mean to ${\approx}35\%$,
which remains $4.5\times$ the rest mean.}
\end{table}

\section{Beta-NLL Intervention Details}
\label{app:beta_nll_details}


We compare four MAP-based variants on Kin8nm to separate possible
sources of the observed instability.
MAP5 and MAP10 train five and ten independent runs per realization and
select the best by validation CRPS, testing whether unlucky
initialization drives the effect.
The long-training variant extends optimization to 2000 epochs, testing
whether the baseline simply needs more iterations.
MAP+$\beta$-NLL \citep{seitzer2022pitfalls,stirn2023faithful} replaces the standard objective with $\beta$-NLL
($\beta=0.5$), testing whether the instability is linked to the
heteroscedastic training objective itself.

Results are summarized in Table~\ref{tab:beta_nll_signatures}.
Additional restarts and longer training improve $R^2$ only partially
and leave the trajectory non-monotone.
$\beta$-NLL substantially restores regular variance contraction on all
three real datasets where the effect is tested.

\begin{table}[h]
\centering
\caption{Stability signatures for MAP-based intervention variants.}
\label{tab:beta_nll_signatures}
\begin{tabular}{llcc}
\toprule
Dataset & Method & $R^2$ & Monotone \\
\midrule
\multirow{5}{*}{Kin8nm}
  & MAP (baseline)   & $0.158$ & No  \\
  & MAP5             & $0.543$ & No  \\
  & MAP10            & $0.583$ & No  \\
  & MAP (2000 ep.)   & $0.511$ & No  \\
  & MAP+$\beta$-NLL  & $0.948$ & Yes \\
\midrule
\multirow{2}{*}{Tetouan}
  & MAP (baseline)   & $0.825$ & No  \\
  & MAP+$\beta$-NLL  & $0.989$ & Yes \\
\midrule
\multirow{2}{*}{Seoul Bike}
  & MAP (baseline)   & $0.530$ & No  \\
  & MAP+$\beta$-NLL  & $0.976$ & Yes \\
\bottomrule
\end{tabular}
\end{table}

\section{Additional Empirical Results}
\label{app:add_results}

\subsection{Robustness to dataset and method dependence}
\label{app:fe_robustness}

As a robustness check, we fit a fixed-effects linear model with log relative RMSE as the response, log local variance as the predictor, and dataset and method indicators as controls. The coefficient of log local variance remains positive and highly significant after accounting for these average differences across datasets and methods (\(\hat\beta = 0.346\), SE \(= 0.009\), 95\% CI \([0.329, 0.363]\)). This indicates that the variance--error relationship is not an artifact of pooling heterogeneous method--dataset combinations.

The same pattern also appears within each real dataset. Dataset-specific slopes from separate log--log regressions are all positive, ranging from \(0.274\) on Make Regression to \(0.401\) on Protein. This supports the conclusion that larger local variance is consistently associated with larger single-seed estimation error across datasets.

\begin{table}[h]
\centering
\caption{Fixed-effects robustness check for the variance--error relationship.}
\label{tab:fe_robustness}
\begin{tabular}{lcc}
\toprule
Model / Dataset & Slope on log local variance & Significance \\
\midrule
Fixed effects with dataset and method controls & 0.346 & \(p < 10^{-10}\) \\
Kin8nm & 0.315 & \(p < 10^{-10}\) \\
Seoul Bike & 0.325 & \(p < 10^{-10}\) \\
Make Regression & 0.274 & \(p < 10^{-10}\) \\
Protein & 0.401 & \(p < 10^{-10}\) \\
Tetouan & 0.380 & \(p < 10^{-10}\) \\
\bottomrule
\end{tabular}
\end{table}

\subsection{BBB on Protein: episodic divergence}
\label{app:bbb_protein}

\begin{figure}[h]
\centering
\includegraphics[width=0.72\linewidth]{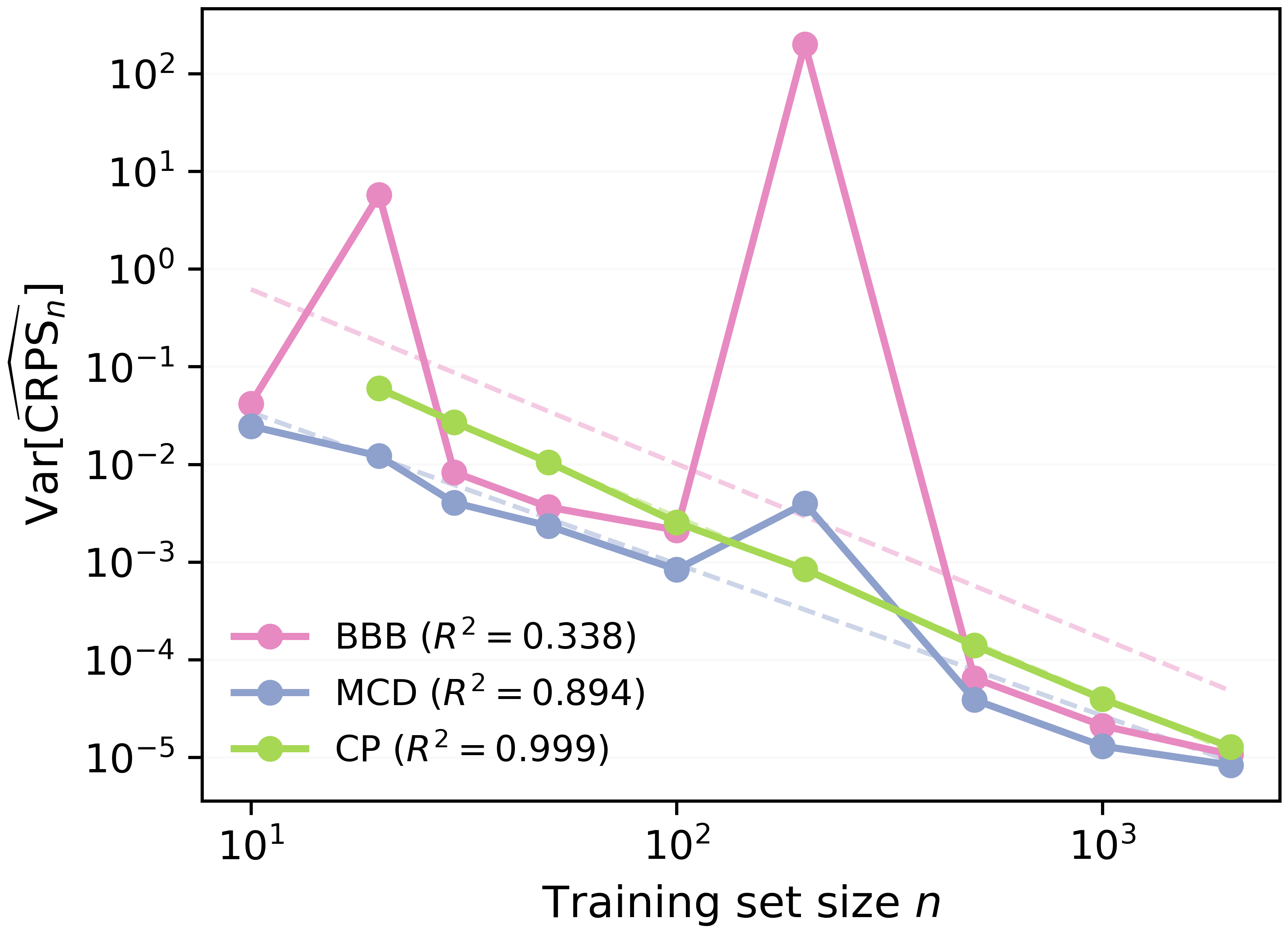}
\caption{Variance trajectory on Protein for BBB with MCD shown as a reference.
BBB exhibits two extreme spikes at $n=20$ and $n=200$, which dominate the
power-law fit and drive the low overall $R^2$. Outside these two divergence
events, the trajectory is much more regular.}
\label{fig:bbb_protein}
\end{figure}

BBB achieves $R^2=0.338$ on Protein because its variance trajectory contains
two extreme spikes at $n=20$ and $n=200$, far above the surrounding values.
These events are better understood as episodic divergence on a small number of
runs than as a broad failure of regular scaling. Figure~\ref{fig:bbb_protein}
shows that the trajectory is otherwise much more regular, and
Table~\ref{tab:bbb_excluding} confirms this interpretation quantitatively:
excluding the two divergence points improves the fitted $R^2$ from $0.338$ to
$0.979$.

\begin{table}[h]
\centering
\caption{BBB on Protein: power-law fit before and after excluding
the two divergence events.}
\label{tab:bbb_excluding}
\begin{tabular}{lcc}
\toprule
Condition & $\alpha$ & $R^2$ \\
\midrule
All data                     & $1.787$ & $0.338$ \\
Excluding $n=20$ and $n=200$ & $1.650$ & $0.979$ \\
\bottomrule
\end{tabular}
\end{table}

\subsection{Full CRPS mean and standard deviation}
\label{app:full_results}

\begin{figure}[h]
\centering
\includegraphics[width=\linewidth]{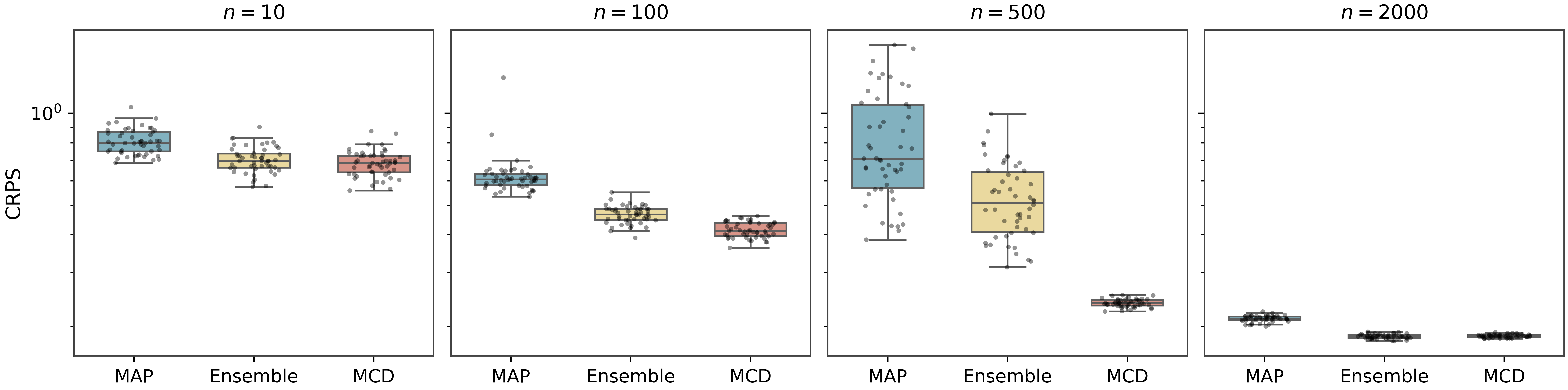}
\caption{Distribution of CRPS across $R=50$ realizations at selected
training sizes on Kin8nm.}
\label{fig:crps_boxplot}
\end{figure}

Tables~\ref{tab:crps_smalln} and~\ref{tab:crps_largen} report
mean $\pm$ std of CRPS across $R=50$ realizations.
Figure~\ref{fig:crps_boxplot} shows CRPS distributions at selected
training sizes; the spread reflects the variability that underlies
the variance trajectories in the main text.

\begin{table*}[t]
\centering
\small
\setlength{\tabcolsep}{4pt}
\caption{CRPS mean $\pm$ std at smaller training sizes.}
\label{tab:crps_smalln}
\begin{tabular}{llrrrrr}
\toprule
Dataset & Method & 10 & 20 & 30 & 50 & 100 \\
\midrule
\multirow{4}{*}{Synthetic}
& MAP      & $0.50\pm0.11$ & $0.35\pm0.06$ & $0.31\pm0.05$ & $0.28\pm0.03$ & $0.26\pm0.02$ \\
& MCD      & $0.50\pm0.11$ & $0.32\pm0.06$ & $0.26\pm0.05$ & $0.21\pm0.03$ & $0.17\pm0.01$ \\
& Ensemble & $0.43\pm0.11$ & $0.26\pm0.05$ & $0.23\pm0.03$ & $0.21\pm0.03$ & $0.19\pm0.01$ \\
& BBB      & $0.44\pm0.10$ & $0.28\pm0.05$ & $0.23\pm0.03$ & $0.20\pm0.02$ & $0.18\pm0.01$ \\
\midrule
\multirow{4}{*}{Kin8nm}
& MAP      & $0.81\pm0.08$ & $0.77\pm0.06$ & $0.76\pm0.06$ & $0.70\pm0.05$ & $0.62\pm0.11$ \\
& MCD      & $0.69\pm0.07$ & $0.61\pm0.06$ & $0.57\pm0.04$ & $0.50\pm0.03$ & $0.42\pm0.02$ \\
& Ensemble & $0.71\pm0.07$ & $0.64\pm0.06$ & $0.61\pm0.06$ & $0.56\pm0.04$ & $0.47\pm0.03$ \\
& BBB      & $0.74\pm0.07$ & $0.68\pm0.05$ & $0.64\pm0.04$ & $0.58\pm0.04$ & $0.48\pm0.03$ \\
\midrule
\multirow{4}{*}{Protein}
& MAP      & $1.17\pm0.31$ & $1.05\pm0.25$ & $0.97\pm0.17$ & $0.86\pm0.08$ & $0.80\pm0.19$ \\
& MCD      & $0.88\pm0.16$ & $0.75\pm0.11$ & $0.68\pm0.06$ & $0.61\pm0.05$ & $0.53\pm0.03$ \\
& Ensemble & $1.03\pm0.29$ & $0.98\pm0.39$ & $0.88\pm0.16$ & $0.86\pm0.24$ & $0.77\pm0.23$ \\
& BBB      & $0.99\pm0.20$ & $1.16\pm2.39$ & $0.76\pm0.09$ & $0.68\pm0.06$ & $0.58\pm0.05$ \\
\midrule
\multirow{4}{*}{Make}
& MAP      & $0.51\pm0.11$ & $0.32\pm0.06$ & $0.22\pm0.04$ & $0.17\pm0.02$ & $0.12\pm0.01$ \\
& MCD      & $0.46\pm0.11$ & $0.31\pm0.08$ & $0.21\pm0.05$ & $0.13\pm0.02$ & $0.08\pm0.01$ \\
& Ensemble & $0.41\pm0.10$ & $0.23\pm0.06$ & $0.15\pm0.03$ & $0.10\pm0.02$ & $0.07\pm0.00$ \\
& BBB      & $0.50\pm0.10$ & $0.31\pm0.08$ & $0.19\pm0.04$ & $0.11\pm0.02$ & $0.07\pm0.01$ \\
\midrule
\multirow{4}{*}{Tetouan}
& MAP      & $1.12\pm0.42$ & $1.30\pm1.03$ & $1.47\pm0.79$ & $1.45\pm1.12$ & $1.05\pm0.60$ \\
& MCD      & $0.78\pm0.14$ & $0.72\pm0.13$ & $0.67\pm0.10$ & $0.60\pm0.05$ & $0.53\pm0.04$ \\
& Ensemble & $1.09\pm0.59$ & $1.24\pm1.01$ & $1.42\pm0.83$ & $1.51\pm0.92$ & $1.03\pm0.48$ \\
& BBB      & $0.91\pm0.22$ & $0.86\pm0.16$ & $0.79\pm0.12$ & $0.71\pm0.09$ & $0.60\pm0.06$ \\
\midrule
\multirow{4}{*}{Seoul Bike}
& MAP      & $0.84\pm0.13$ & $0.83\pm0.12$ & $0.84\pm0.22$ & $0.87\pm0.36$ & $0.95\pm0.48$ \\
& MCD      & $0.71\pm0.10$ & $0.65\pm0.09$ & $0.61\pm0.06$ & $0.56\pm0.04$ & $0.51\pm0.03$ \\
& Ensemble & $0.74\pm0.10$ & $0.74\pm0.11$ & $0.74\pm0.17$ & $0.87\pm0.45$ & $1.11\pm0.56$ \\
& BBB      & $0.74\pm0.11$ & $0.72\pm0.10$ & $0.68\pm0.08$ & $0.62\pm0.06$ & $0.55\pm0.04$ \\
\bottomrule
\end{tabular}
\end{table*}

\begin{table*}[t]
\centering
\small
\setlength{\tabcolsep}{4pt}
\caption{CRPS mean $\pm$ std at larger training sizes.}
\label{tab:crps_largen}
\begin{tabular}{llrrrr}
\toprule
Dataset & Method & 200 & 500 & 1000 & 2000 \\
\midrule
\multirow{4}{*}{Synthetic}
& MAP      & $0.24\pm0.01$ & $0.21\pm0.01$ & $0.17\pm0.00$ & $0.15\pm0.00$ \\
& MCD      & $0.15\pm0.01$ & $0.14\pm0.00$ & $0.13\pm0.00$ & $0.13\pm0.00$ \\
& Ensemble & $0.18\pm0.01$ & $0.16\pm0.00$ & $0.15\pm0.00$ & $0.13\pm0.00$ \\
& BBB      & $0.16\pm0.01$ & $0.14\pm0.00$ & $0.13\pm0.00$ & $0.13\pm0.00$ \\
\midrule
\multirow{4}{*}{Kin8nm}
& MAP      & $0.65\pm0.22$ & $0.83\pm0.34$ & $0.27\pm0.01$ & $0.21\pm0.01$ \\
& MCD      & $0.32\pm0.02$ & $0.24\pm0.01$ & $0.21\pm0.00$ & $0.19\pm0.00$ \\
& Ensemble & $0.47\pm0.11$ & $0.53\pm0.15$ & $0.23\pm0.01$ & $0.19\pm0.00$ \\
& BBB      & $0.37\pm0.02$ & $0.26\pm0.01$ & $0.22\pm0.01$ & $0.20\pm0.00$ \\
\midrule
\multirow{4}{*}{Protein}
& MAP      & $0.64\pm0.09$ & $0.49\pm0.01$ & $0.44\pm0.01$ & $0.42\pm0.00$ \\
& MCD      & $0.50\pm0.06$ & $0.44\pm0.01$ & $0.43\pm0.00$ & $0.42\pm0.00$ \\
& Ensemble & $0.60\pm0.11$ & $0.47\pm0.01$ & $0.43\pm0.01$ & $0.41\pm0.00$ \\
& BBB      & $2.50\pm14.09$ & $0.45\pm0.01$ & $0.43\pm0.00$ & $0.43\pm0.00$ \\
\midrule
\multirow{4}{*}{Make}
& MAP      & $0.09\pm0.01$ & $0.07\pm0.01$ & $0.06\pm0.01$ & $0.05\pm0.01$ \\
& MCD      & $0.06\pm0.00$ & $0.05\pm0.00$ & $0.05\pm0.00$ & $0.04\pm0.00$ \\
& Ensemble & $0.06\pm0.00$ & $0.05\pm0.00$ & $0.04\pm0.00$ & $0.04\pm0.00$ \\
& BBB      & $0.05\pm0.00$ & $0.05\pm0.00$ & $0.05\pm0.00$ & $0.05\pm0.00$ \\
\midrule
\multirow{4}{*}{Tetouan}
& MAP      & $0.62\pm0.05$ & $0.51\pm0.02$ & $0.48\pm0.01$ & $0.47\pm0.01$ \\
& MCD      & $0.49\pm0.02$ & $0.46\pm0.01$ & $0.45\pm0.01$ & $0.45\pm0.01$ \\
& Ensemble & $0.59\pm0.06$ & $0.50\pm0.02$ & $0.47\pm0.01$ & $0.46\pm0.01$ \\
& BBB      & $0.53\pm0.03$ & $0.48\pm0.02$ & $0.46\pm0.01$ & $0.46\pm0.01$ \\
\midrule
\multirow{4}{*}{Seoul Bike}
& MAP      & $0.78\pm0.37$ & $0.49\pm0.02$ & $0.45\pm0.01$ & $0.43\pm0.01$ \\
& MCD      & $0.46\pm0.02$ & $0.44\pm0.01$ & $0.42\pm0.01$ & $0.42\pm0.00$ \\
& Ensemble & $0.94\pm0.49$ & $0.45\pm0.01$ & $0.43\pm0.01$ & $0.42\pm0.01$ \\
& BBB      & $0.49\pm0.02$ & $0.44\pm0.01$ & $0.43\pm0.01$ & $0.42\pm0.00$ \\
\bottomrule
\end{tabular}
\end{table*}

\subsection{Scaling exponents}
\label{app:scaling_exponents}

The scaling exponent $\alpha$ alone does not fully characterize the
variance trajectory: non-monotone behavior cannot be captured without
also considering $R^2$.
The $(\alpha, R^2)$ pairs are provided in Table~\ref{tab:powerlaw}
for completeness; the main text focuses on trajectory shape and fit
quality rather than the exponent value.

\begin{table}[h]
\centering
\small
\setlength{\tabcolsep}{3pt}
\caption{Power-law fit results ($\alpha$ and $R^2$) for each
method and dataset.}
\label{tab:powerlaw}
\begin{tabular}{llcccccccccccc}
\toprule
& & \multicolumn{2}{c}{Synthetic}
  & \multicolumn{2}{c}{Kin8nm}
  & \multicolumn{2}{c}{Protein}
  & \multicolumn{2}{c}{Make}
  & \multicolumn{2}{c}{Tetouan}
  & \multicolumn{2}{c}{Bike} \\
\cmidrule(lr){3-4}\cmidrule(lr){5-6}\cmidrule(lr){7-8}
\cmidrule(lr){9-10}\cmidrule(lr){11-12}\cmidrule(lr){13-14}
Type & Method
  & $\alpha$ & $R^2$
  & $\alpha$ & $R^2$
  & $\alpha$ & $R^2$
  & $\alpha$ & $R^2$
  & $\alpha$ & $R^2$
  & $\alpha$ & $R^2$ \\
\midrule
\multirow{2}{*}{Direct}
& MAP
  & 1.40 & 1.00 & 0.56 & 0.16
  & 1.67 & 0.88 & 1.06 & 0.84
  & 1.96 & 0.83 & 1.29 & 0.53 \\
& Ensemble
  & 1.53 & 0.98 & 0.80 & 0.34
  & 1.92 & 0.87 & 1.49 & 0.88
  & 2.05 & 0.88 & 1.34 & 0.45 \\
\midrule
\multirow{2}{*}{Implicit}
& MCD
  & 1.91 & 1.00 & 1.30 & 0.97
  & 1.55 & 0.89 & 2.29 & 0.98
  & 1.23 & 0.99 & 1.19 & 0.99 \\
& BBB
  & 1.65 & 0.99 & 1.10 & 0.98
  & 1.79 & 0.34 & 2.18 & 0.96
  & 1.30 & 1.00 & 1.26 & 0.98 \\
\midrule
\multicolumn{2}{l}{Implicit mean $R^2$}
  & \multicolumn{2}{c}{0.992}
  & \multicolumn{2}{c}{0.972}
  & \multicolumn{2}{c}{0.616}
  & \multicolumn{2}{c}{0.967}
  & \multicolumn{2}{c}{0.992}
  & \multicolumn{2}{c}{0.985} \\
\multicolumn{2}{l}{Direct mean $R^2$}
  & \multicolumn{2}{c}{0.991}
  & \multicolumn{2}{c}{0.247}
  & \multicolumn{2}{c}{0.875}
  & \multicolumn{2}{c}{0.858}
  & \multicolumn{2}{c}{0.852}
  & \multicolumn{2}{c}{0.492} \\
\bottomrule
\end{tabular}
\end{table}

\subsection{Alternative metrics in regression}
\label{app:other_metrics}

Our main analysis uses CRPS because it provides the clearest view of benchmark measurement failure in our setting. To examine whether the same pattern appears under other regression metrics, we repeated the variance--error analysis for negative log likelihood, interval score \citep{bracher2021evaluating}, and interval coverage probability.

Among the alternative metrics, interval score behaves most similarly to CRPS. Across all five real datasets, local variance remains strongly associated with single-seed relative RMSE, with Spearman correlations ranging from \(0.803\) to \(0.992\). The high-variance region also has substantially larger single-seed error than the remaining points on every dataset, and relative RMSE increases monotonically across variance quartiles in all five cases. This indicates that the measurement-failure pattern emphasized in the main text is not unique to CRPS, although it is somewhat weaker under interval score.

By contrast, the same pattern is much less consistent under negative log likelihood. While the variance--error association remains visible on some datasets, especially Protein and Tetouan, it is substantially weaker on Kin8nm and Seoul Bike, and the quartile trend is not uniformly monotone. Negative log likelihood therefore provides a less reliable basis for the diagnostic developed in the main text.

Interval coverage probability provides the weakest signal. Its variance is only weakly associated with single-seed relative RMSE on several datasets, and neither the high-variance comparison nor the quartile analysis yields a consistent pattern. In our setting, PICP therefore does not support the same local-variance diagnostic.

Taken together, these results suggest a clear ordering across regression metrics: CRPS provides the strongest and most consistent view of measurement failure, interval score shows a similar but weaker pattern, negative log likelihood is mixed, and PICP is largely uninformative for this purpose.

\begin{table}[t]
\centering
\small
\setlength{\tabcolsep}{5pt}
\caption{Summary of variance--error diagnostics for alternative regression metrics across the five real datasets. `High-var significant' counts datasets where the top-quartile local-variance region has significantly larger relative RMSE than the remaining points under a one-sided Mann--Whitney test. `Quartile monotone' counts datasets where mean relative RMSE increases monotonically across variance quartiles.}
\label{tab:alt_metrics_summary}
\begin{tabular}{lcccc}
\toprule
Metric & Spearman \(\rho\) range & High-var significant & Quartile monotone & Overall assessment \\
\midrule
CRPS           & \(0.966\) -- \(0.998\) & \(5/5\) & \(5/5\) & Strongest and most consistent \\
Interval score & \(0.803\) -- \(0.992\) & \(5/5\) & \(5/5\) & Similar pattern, weaker than CRPS \\
NLL            & \(0.451\) -- \(0.854\) & \(3/5\) & \(2/5\) & Mixed and dataset-dependent \\
PICP           & \(-0.161\) -- \(0.850\) & \(2/5\) & \(0/5\) & Weak and inconsistent \\
\bottomrule
\end{tabular}
\end{table}

\subsection{Brier score in classification}
\label{app:classification}

We repeated the variance trajectory and reliability analysis on four
binary classification datasets using Brier score \citep{brier1950verification}.
Across all four datasets and all methods, variance trajectories are
monotone and well described by a power-law fit, with no
intermediate-$n$ spikes and no family-level split between direct and
implicit methods.

The local-variance diagnostic holds as strongly as in the regression
setting.
Table~\ref{tab:classification} reports the Spearman correlation
between local $\mathrm{Var}[\widehat{\mathrm{Brier}}_n]$ and
single-seed relative RMSE, together with the mean relative RMSE in
the top-quartile variance region versus the remaining points.
The Spearman correlation exceeds $0.98$ on all four datasets, and
relative RMSE increases monotonically across variance quartiles on
every dataset without exception.

The absence of intermediate-$n$ spikes is itself informative.
Brier score does not involve a heteroscedastic variance head, so the
gradient-attenuation mechanism discussed in
Appendix~\ref{app:mechanism} does not apply.
This contrast supports the view that the intermediate-$n$ instability
in the regression setting is linked to the heteroscedastic NLL
objective rather than being a general property of limited-data
training.

\begin{table}[h]
\centering
\small
\caption{Local-variance diagnostic for binary classification datasets
under Brier score.
Spearman $\rho$ is computed across all method-$n$ pairs.
High-var and rest report mean relative RMSE in the top-quartile and
remaining variance regions respectively.
Quartile monotone indicates whether mean relative RMSE increases
monotonically from Q1 to Q4.
All four datasets also show monotone variance trajectories with
no intermediate-$n$ spikes.}
\label{tab:classification}
\begin{tabular}{lrrrr}
\toprule
Dataset & Spearman $\rho$ & $p$-value & High-var / Rest & Q-mono \\
\midrule
MAGIC          & $0.995$ & $2.1\times10^{-35}$ & $18.3\%$ / $7.1\%$ & Yes \\
Default Credit & $0.996$ & $5.0\times10^{-38}$ & $16.6\%$ / $5.9\%$ & Yes \\
Adult          & $0.997$ & $1.4\times10^{-38}$ & $14.2\%$ / $6.5\%$ & Yes \\
MiniBoone      & $0.989$ & $9.2\times10^{-30}$ & $18.1\%$ / $7.9\%$ & Yes \\
\bottomrule
\end{tabular}
\end{table}

\end{document}